\pdfoutput=1
\documentclass[11pt]{article}

\usepackage[preprint]{acl}

\usepackage{times}
\usepackage{latexsym}

\usepackage[T1]{fontenc}
\usepackage[utf8]{inputenc}

\usepackage{microtype}

\usepackage{inconsolata}

\usepackage{graphicx}
\usepackage{hyperref}
\usepackage{natbib}
\usepackage{amsfonts}
\usepackage{amsmath}
\usepackage{amssymb}
\usepackage{booktabs}
\usepackage{multirow}
\usepackage{xcolor}
\usepackage{colortbl}
\usepackage{adjustbox}
\usepackage{makecell}
\usepackage{subcaption}
\usepackage{caption}
\usepackage{enumitem}

\title{HyperLoader: Integrating Hypernetwork-Based LoRA and Adapter Layers into Multi-Task Transformers for Sequence Labelling}

\author{
  \textbf{Jesus-German Ortiz-Barajas\textsuperscript{1}},
  \textbf{Helena Gómez-Adorno\textsuperscript{2}},
  \textbf{Thamar Solorio\textsuperscript{1}}
\\
  \textsuperscript{1}MBZUAI,
  \textsuperscript{2}IIMAS
\\
\texttt{\{jesus.ortizbarajas, thamar.solorio\}@mbzuai.ac.ae}\\
\texttt{helena.gomez@iimas.unam.mx}
}

\begin{document}
\maketitle
\begin{abstract}
We present HyperLoader, a simple approach that combines different parameter-efficient fine-tuning methods in a multi-task setting. To achieve this goal, our model uses a hypernetwork to generate the weights of these modules based on the task, the transformer layer, and its position within this layer. Our method combines the benefits of multi-task learning by capturing the structure of all tasks while reducing the task interference problem by encapsulating the task-specific knowledge in the generated weights and the benefits of combining different parameter-efficient methods to outperform full-fine tuning. We provide empirical evidence that HyperLoader outperforms previous approaches in most datasets and obtains the best average performance across tasks in high-resource and low-resource scenarios.  
\end{abstract}

\section{Introduction}
Parameter-efficient fine-tuning techniques emerge as an alternative to conventional fine-tuning, where only a small number of parameters is updated to a downstream task \cite{houlsby19a, stickland19a, NEURIPS2021_081be9fd}. These methods aim to achieve comparable performance to full fine-tuning by updating as few parameters as possible. However, a less studied research direction related to these methods is whether one can perform better than full fine-tuning with fewer parameters \cite{mao-etal-2022-unipelt}.

Although these methods considerably reduce the number of parameters to achieve good performance on different tasks, a specialized model for each task remains necessary. Multi-task learning reduces the computational cost by training a single model while enabling information sharing by capturing the common structure underlying all tasks. Recently, researchers have developed approaches combining parameter-efficient fine-tuning techniques in multi-task settings \cite{stickland19a, pfeiffer-etal-2020-mad, pfeiffer-etal-2021-adapterfusion, ruckle-etal-2021-adapterdrop}. Nevertheless, this approach can lead to underfitting in high-resource tasks and overfitting in low-resource tasks \cite{lee-etal-2017-fully}. Another potential issue is task interference, where an improvement in the performance of one task reduces the performance of other tasks \cite{Wang_2019_CVPR}.

An alternative to reduce the negative effects of multi-task learning is to use hypernetworks \cite{ha2016hypernetworks} to generate separate weights for each task \cite{karimi-mahabadi-etal-2021-parameter, ivison-peters-2022-hyperdecoders}. However, combining different parameter-efficient methods in a multi-task setting using hypernetworks is an unexplored research area. To address this limitation, we propose HyperLoader, a simple method that employs a neural network to generate the weights for a combination of two parameter-efficient fine-tuning techniques: adapters \cite{houlsby19a} and LoRA \cite{hu2021lora} based on the task, the transformer layer of the model and the position of the method within this layer. We use the encoder-decoder T5 model \cite{raffel2020exploring} for all experiments to take advantage of modelling the tasks as sequence-to-sequence tasks. We test our model in seven datasets from two Sequence Labelling tasks. The first task is Named Entity Recognition, a valuable tool in various real-world scenarios in the era of large language models such as healthcare and medical research \cite{raza2022large, 10.1093/jamia/ocad259}, Finance and Business Intelligence \cite{zhou2023universalner}, and analyzing legal texts \cite{trias-etal-2021-named}. The second task is slot-filling, a crucial step to enable dialogue systems to accurately understand and fulfil user requests by extracting and organizing essential information from user inputs \cite{cheng-etal-2023-mrrl, cheng-etal-2023-accelerating, firdaus2023multitask}.   

Our model achieves the best average performance using the complete training and validation data for each task, as well as in a low-resource configuration with only 10\% and 20\% of the training and validation data available. We empirically demonstrate that the improvement is not only due to adding more trainable parameters but also by combining different parameter-efficient methods in a multi-task setting using hypernetworks, which adequately supports Sequence Labelling tasks in high- and low-resource scenarios.

Our main contributions are the following:
\begin{itemize}[noitemsep]
    \item We propose a multi-task learning approach that combines different parameter-efficient fine-tuning methods based on hypernetworks conditioned on the task, the layer in the transformer model, and the position of the method within this layer.
    \item We provide empirical results on different Sequence Labelling datasets demonstrating the effectiveness of our model compared to single-task and multi-task approaches using the entire datasets and in low-resource scenarios.   
\end{itemize}

\section{Related Work}\label{sec:related work}
Parameter-efficient fine-tuning techniques are an alternative to full fine-tuning by updating a small number of parameters per task, yielding similar performance to full fine-tuning. One of the most popular approaches in this area is the adapter \cite{houlsby19a}, a small trainable bottleneck layer added in each transformer block of a model. It consists of down and up projections and has shown similar performance to full-fine-tuning. Other popular parameter-efficient fine-tuning approaches are Prefix-tuning \cite{li-liang-2021-prefix} and LoRA \cite{hu2021lora}. Prefix-tuning prepends a fixed-length, learnable sequence of prefix tokens to the input embeddings while the original model parameters remain unchanged. This approach optimizes a continuous prompt, effectively guiding the model's behaviour for specific tasks with minimal computational overhead. LoRA adds trainable low-rank matrices and combines their outputs with the original matrices in the self-attention layer of the transformer; the main difference with the previous approaches is that it does not employ any activation function.

A less-studied area is how to achieve better performance than full-fine tuning using as few parameters as possible. The UniPELT framework \cite{mao-etal-2022-unipelt} addresses this limitation. This framework incorporates several parameter-efficient fine-tuning methods as sub-modules and learns how to activate them dynamically depending on the input or task setup using a gating mechanism, which assigns more weight to the sub-modules that increase the performance of a given task. They considered three different methods: Adapters, LoRA, and prefix-tuning, and they used the BERT base model for all their experiments.

Although promising, this approach still needs a specialized model for each task. The use of hypernetworks to generate the weights of parameter-efficient methods in a multi-task learning approach is a promising research area. By using this type of network, the model can capture the common structure underlying all target tasks and encapsulate the specific task knowledge by generating different weights based on the task, reducing the negative effects of this setting such as underfitting and overfitting in high-resource and low-resource tasks, respectively or task interference.

The HyperFormer++ model \cite{karimi-mahabadi-etal-2021-parameter} is the main foundation of our model. It uses task-conditioned hypernetworks to generate the weights of adapter layers and layer normalizations in a multi-task setting. The hypernetwork is trained to create adapter parameters specific to each task and layer, based on the embeddings of task and layer IDs. This hypernetwork is trained simultaneously for all tasks, allowing it to share knowledge across different tasks. At the same time, it reduces negative interference by producing distinct adapter layers for each task. The T5 model is used as the backbone model for all the experiments.

Another hypernetwork-based multi-task approach is Hyperdecoders \cite{ivison-peters-2022-hyperdecoders}. This framework generates input-conditioned adapter layers for the decoder in an encoder-decoder model. This approach offers more flexibility by generating unique parameters for every input instead of using a learnt task embedding and allows making use of the similarities between samples across datasets.

Our HyperLoader model combines the benefits of both worlds. It uses different parameter-efficient fine-tuning techniques and task-conditional hypernetworks to generate its weights conditioned on the task, the transformer layer of the model, and the position of the parameter-efficient method within this layer. Our results show a performance improvement in all tasks and, on average, using the full training and validation data and simulating a low-resource setting.    

\section{Methodology}\label{sec:methodology}
Multi-task hypernetwork-based approaches using parameter-efficient methods are a low-cost alternative for different NLP tasks that achieve comparable results to full-fine-tuning. Nevertheless, exploring approaches to obtain better results than updating all weights during fine-tuning is an unexplored research area. We introduce the HyperLoader model to address this gap. Our approach incorporates \textbf{hyper}network-based \textbf{Lo}RA and \textbf{Ad}apter layers into a multi-task transform\textbf{er} model, based on the HyperFormer++ model \cite{karimi-mahabadi-etal-2021-parameter}.

\begin{figure}
    \centering
    \includegraphics[scale=0.225]{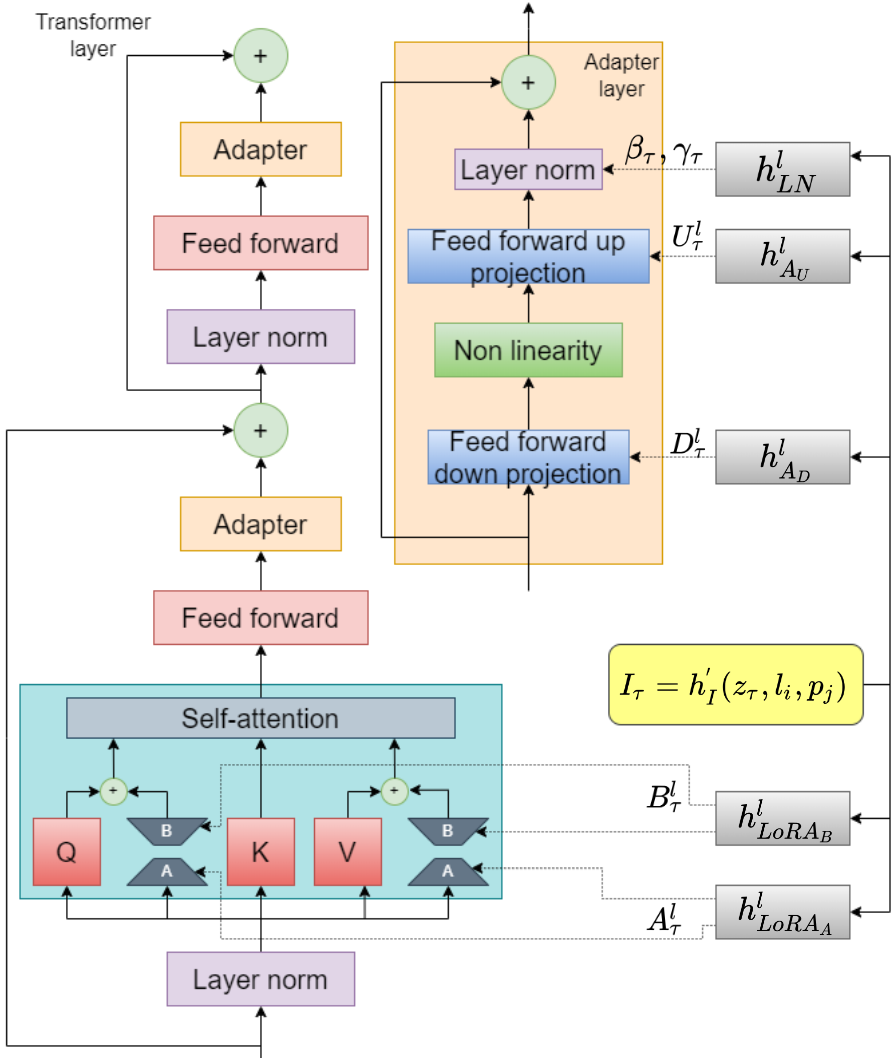}
    \caption{Diagram of the HyperLoader model. The Adapter hypernetworks $h^{l}_{A_D}$ and $h^{l}_{A_U}$ produce the weights $D^{l}_{\tau}$ and $U^{l}_{\tau}$ for task-specific adapter modules. The LoRA hypernetworks $h^{l}_{LoRA_{A}}$ and $h^{l}_{LoRA_{B}}$ generate the $A$ and $B$ matrices for task-specific LoRA modules. Finally, the hypernetwork $h^{l}_{LN}$ creates the conditional layer normalization parameters $\beta_{\tau}$ and $\gamma_{\tau}$.}
    \label{fig:hyperloader}
\end{figure}

Our method enables the hypernetwork to generate the weights of the adapter layers and LoRA matrices based on the task, the transformer model's layer, and the position of the adapter and LoRA matrix within this layer. Figure \ref{fig:hyperloader} shows the diagram of our proposed approach. We generate an input embedding $I_{\tau}$ based on the task $\tau$, the layer $l_i$ and the position of the Adapter and the LoRA matrices $p_j$. The embedding $I_{\tau}$ is the input to a set of hypernetworks ($h_{LN}^l$, $h^l_{A_U}$, $h^l_{A_D}$, $h^l_{LoRA_B}$, $h^l_{LoRA_A}$) to generate the weights for the adapter layers, the LoRA matrices and the layer normalization. Therefore, we only update the hypernetwork parameters, the input embedding projector $h^{'}$ and the layer normalizations in $f_\theta(\cdot)$ during training, while the rest of parameters $\theta$ remain fixed.  

In this section, we describe our approach. First, we provide preliminary information by defining multi-task learning, which is the foundation of our work. Second, we describe the main components of our proposed HyperLoader model. Finally, we describe the format for converting Sequence Labelling into a sequence-to-sequence task. 

\subsection{Preliminaries: Multi-task Learning}
Our proposed model generates the weights of adapter layers and LoRA matrices based on the task, the transformer layer of the model, and the position of the adapter and the LoRA matrix in this layer in a multi-task setting. Given a set of tasks $\{D_{{\tau}_{\tau=1}^{T}\}}$ where $T$ is the total number of tasks, $D_{\tau}=\{(x_\tau^i, y_\tau^i)\}_{i=1}^{N_\tau}$ is the training data of the $\tau$-th task with $N_\tau$ instances and $f_{\theta}(\cdot)$ is a pre-trained model parameterized by $\theta$, the multi-task fine-tuning minimizes the loss shown in equation \ref{eq:multi_task_loss} on the training set, where $l$ is the cross-entropy loss and $w_\tau$ is the sampling weight for the $\tau$-th task.

{\small
\begin{equation}
    \mathcal{L}(\theta, \{ D_{\tau}\}_{\tau=1}^T)=\sum_{\tau=1}^{T} \sum_{(x_{\tau}^i , y_{\tau}^i) \in D_{\tau}}w_{\tau}l(f_{\theta}(x_{\tau}^i),y_{\tau}^i)
    \label{eq:multi_task_loss}
\end{equation}}

\subsection{HyperLoader: Task Conditional Adapter Layers}
Adapters \cite{houlsby19a} are small sub-modules inserted within layers of a pre-trained transformer-based model before the skip connections. In the HyperLoader model, the conditional adapter $A^l$ for layer $l$ consists of a down-projection $\mathcal{D}^l_{\tau} \in \mathbb{R}^{h \times d}$ into a lower dimension $d_{bottleneck}$, a $GeLU$ non-linearity, and an up-projection $\mathcal{U}^l_{\tau} \in \mathbb{R}^{d \times h}$ that projects back into the original hidden layer dimension. The output of the adapter is defined in equation \ref{eq:adapter}, where $x$ is the input hidden state, $LN^l_{\tau}$ is the task conditional layer normalization and the adapter weights $(\mathcal{D}^l_{\tau}, \mathcal{U}^l_{\tau})$ are generated by a hypernetwork. We use a reduction factor $r=32$ for the down-projection in all the experiments.

{\small
\begin{equation}
    A^l_{\tau}(x) = LN^l_{\tau}\left(\mathcal{U}^l_{\tau}(GeLU(\mathcal{D}^l_{\tau}(x)))\right) + x
    \label{eq:adapter}
\end{equation}}

\subsection{HyperLoader: Task Conditional Layer Normalization}
The task conditional layer normalization, defined in equation \ref{eq:layer_norm}, involves element-wise multiplication denoted by $\odot$. Here, $\mu_{\tau}$ and $\sigma_{\tau}$ represent the mean and standard deviation of the training data for the $\tau$-th task, while $\gamma^l_{\tau}$ and $\beta^l_{\tau}$ are weights generated by a hypernetwork. 

{\small
\begin{equation}
     LN^l_{\tau}(x^i_{\tau}) = \gamma^l_{\tau} \odot \frac{x^i_{\tau} - \mu_{\tau}}{\sigma_{\tau}}+\beta^l_{\tau} 
     \label{eq:layer_norm}
\end{equation}}

\subsection{HyperLoader: Task Conditional LoRA Matrices}
Low-Rank Adaptation (LoRA) \cite{hu2021lora} is a parameter-efficient fine-tuning technique that injects trainable rank decomposition matrices into each layer of the Transformer architecture. For a pre-trained weight matrix $W_0 \in \mathbb{R}^{d \times k}$, this method constrains its update by representing it with a low-rank decomposition $W_0 + \Delta W = W_0 + BA$, where $B \in\mathbb{R}^{d\times r}$ and $A \in\mathbb{R}^{r\times k}$. The forward pass using LoRA is denoted in equation \ref{eq:lora_pass}. Where $x$ is the input hidden state, $W_0$ is frozen and does not receive gradient updates and matrices $A$ and $B$ are generated by a hypernetwork. Based on previous work \cite{hu2021lora, mao-etal-2022-unipelt}, we only add Low-Rank matrices to the query and value matrices in the attention layers of the transformer. We use a low-dimensional rank $r=8$.

{\small
\begin{equation}
    h = W_0x + \Delta Wx = W_0x + BAx
    \label{eq:lora_pass}
\end{equation}}

\subsection{HyperLoader: Task Conditional Hypernetworks}
A hypernetwork \cite{ha2016hypernetworks} is a neural network that generates the parameters for another neural network. It captures shared information across tasks, while the generated task-specific adapters and layer normalization allow the model to adapt to each task individually, reducing negative interference in a multi-task setting. The input for these networks is a task embedding $z_{\tau}$ and its concatenation with a layer \textit{id} embedding $\mathcal{I} = \{l_i\}_{i=1}^L$ and an adapter position embedding $\mathcal{P}=\{ p_j \}_{j=1}^6$. In the encoder-decoder model, we consider six positions for the parameter-efficient modules: the adapter layers and the LoRA matrices. We consider two options for the position of the adapter layers in each transformer block: after the self-attention layer or after the feed-forward layer. For LoRA matrices, we consider four positions: the A and B matrices in the self-attention layer and the A and B matrices in the cross-attention layer in the decoder of the model.

Using this configuration, the hypernetwork is able to generate different weights based on the task, position of the adapter, LoRA matrices and layer of the transformer network. The final input embedding $I_{\tau}$ for the hypernetwork is computed using a projector network $h^{'}_{I}$ as shown in equation \ref{eq:input_embedding_hn}, where $z_{\tau}$, $l_i$ and $p_j$ are learnable parameters via back-propagation. We set the dimension of the task feature embedding $z_{\tau}$ to $d_{z_{\tau}} = 512$. For the layer and position embeddings, we use a dimension $d_{l,p} = 64$. Finally, for the final input embedding, we set a dimension of $d_{I_{\tau}} = 64$.

We use five hypernetworks for the HyperLoader model. For the Adapters, we consider two hypernetworks: one for the down projection and one for the up projection. Since we are using a reduction factor $r = 32$ and the hidden dimension of the model is $d=768$, the down sample size is $\lfloor 768/32 \rfloor = 24$, the dimensions of the output matrices of the down and up projections are $768 \times 24$ and $24 \times 768$, respectively. We use two hypernetworks to generate the weights of the LoRA Matrices. One for the low-rank matrix $A$ and one for the low-rank matrix $B$. We use a low-dimensional rank $r=8$ for the Query and Value matrices in the self-attention and the cross-attention; therefore, the output matrices A and B dimensions are $768 \times 8$ and $8 \times 768$, respectively. Finally, we use one hypernetwork for the task-conditioned layer normalization that generates vector weights of $768$ dimensions.

{\small
\begin{equation}
    I_{\tau} = h^{'}_{I}(z_{\tau},l_i,p_j)
    \label{eq:input_embedding_hn}
\end{equation}}

\subsection{SentT' Format}
Converting Sequence Labelling into a sequence-to-sequence task is essential to using an encoder-decoder model. We use the SentT' \cite{farina2023distillation} format for the input and output sequences to achieve this goal. This format is obtained by turning a labelled sequence $S$ into a sequence of tokens $x_1 \dots x_L$ of size $L$ using white-space splitting and then interleaving these tokens with the spacial sentinel tokens used to pre-train T5 to obtain the input string $S_{in} = s_0x_1s1\dots x_Ls_L$. We use the \textit{simplified Beginning, Inside and Outside} (sBIO) format to represent the output strings, where given a set $T$ of labels to annotate a text span, $t \in T$ is used to represent any token that starts a labelled span, a single tag $I$ for each token that continues a labelled span and $O$ to tag tokens that do not belong to labelled spans. Analogous to the input sequence, we interleave the tokenized output string with the sentinel tokens to obtain the output sequence using the SentT' format $S_{out} = s_0t_1s_1\dots t_Ls_L$. Table \ref{tab:SentT_format} illustrates how to convert the text "play the song little robin redbreast" from the SNIPS dataset into the SentT' format.  

\begin{table*}
    \centering
    \resizebox{\textwidth}{!}{%
    \renewcommand{\arraystretch}{0.80}
    \begin{tabular}{lllllll}
    \toprule
    %\multicolumn{6}{c}{\textbf{Original text}}\\
    %\hline
    %play & the & song & little & robin & redbreast & \\
    %\midrule
    \multicolumn{7}{c}{\textbf{Encoded input using the SentT' format}}\\
    \midrule
    \colorbox{lightgray}{<extra\_id\_0>} play &\colorbox{lightgray}{<extra\_id\_1>} the & \colorbox{lightgray}{<extra\_id\_2>} song & \colorbox{lightgray}{<extra\_id\_3>} little &\colorbox{lightgray}{<extra\_id\_4>} robin &\colorbox{lightgray}{<extra\_id\_5>} redbreast & \colorbox{lightgray}{<extra\_id\_6>}\\ 
    \midrule
    \multicolumn{7}{c}{\textbf{Encoded output using the SentT' format}}\\
    \midrule
    \colorbox{lightgray}{<extra\_id\_0>} \textcolor{blue}{O} &\colorbox{lightgray}{<extra\_id\_1>} \textcolor{blue}{O} & \colorbox{lightgray}{<extra\_id\_2>} \textcolor{orange}{MUSIC\_ITEM} & \colorbox{lightgray}{<extra\_id\_3>} \textcolor{orange}{TRACK} & \colorbox{lightgray}{<extra\_id\_4>} \textcolor{orange}{I} &\colorbox{lightgray}{<extra\_id\_5>} \textcolor{orange}{I} & \colorbox{lightgray}{<extra\_id\_6>}\\ 
    \midrule
    \end{tabular}}
    \caption{Example of an input/output instance transformed into the SentT' format. The text is tokenized with white-space splitting and then interleaved with sentinel tokens used for T5 pre-training to obtain the input string. The output uses the sBIO format: a label $t\in T$ is used to represent any token that starts a labelled span, a single tag $I$ for tokens that continue a labelled span and $O$ to tag tokens outside labelled spans.}
    \label{tab:SentT_format}
\end{table*}

\section{Experiments}
This section describes our experiments for Sequence Labelling using the HyperLoader model. %First, we introduce the datasets we used to train and evaluate our approach. Second, we provide the implementation, hyper-parameter and evaluation details of all experiments.    

\subsection{Corpora}
We use seven publicly available corpora for Sequence Labelling tasks (slot-filling and Named Entity Recognition) to cover different domains and distributions and gather robust empirical evidence. 

For slot-filling, we use four different dialogue-oriented datasets. First, the ATIS dataset \cite{hemphill-etal-1990-atis} consists of manual transcripts of audio recordings about people asking for flight information on automated airline travel inquiry systems. Second, the SNIPS dataset \cite{coucke2018snips} comprises users' intent queries distributed in seven domains: search creative work, get weather conditions, restaurant reservations, play music, add elements to a playlist, and search screening events. The third and fourth datasets are the English portions of two multilingual task-oriented dialogue datasets: The mTOP dataset \cite{li-etal-2021-mtop} consists of eleven domains: alarm, calling, event, messaging, music, news, people, recipes, remainders, timer and weather conditions. Finally, the mTOD dataset \cite{schuster-etal-2019-cross-lingual} contains nine labels across three domains: alarm, reminders and weather conditions.

For Named Entity Recognition, we use the MIT Corpora (Movie, MovieTrivia and Restaurant)\footnote{Downloaded from \url{https://groups.csail.mit.edu/sls/}}. In the case of Movie and MovieTrivia datasets, both are composed of 12 labels, and the latter contains more complex queries. Finally, the Restaurant dataset contains 8 labels such as restaurant names, ratings, dishes and opening times. We show the complete list of labels for each dataset in Appendix \ref{sec:appendix}. Based on the label's name, we obtained the label overlap across the datasets. Movie and MovieTrivia have the highest number of overlaps with four labels: genre, year, plot, director, and actor. These datasets also share the labels "genre" and "year" with the SNIPS dataset. The mTOD, mTOP and restaurant datasets share the label "location". Movie and Restaurant datasets share the label "rating". Finally, Restaurant and SNIPS datasets share two labels: "restaurant\_name" and "cuisine". Nine different labels are shared across the datasets, representing 3.78\% of the total of labels.

We used the original partitions for the mTOP and mTOD datasets, in the case of the ATIS and SNIPS datasets, we used the data partitions used originally by \citep{goo-etal-2018-slot}\footnote{Downloaded from \url{https://github.com/MiuLab/SlotGated-SLU/tree/master/data}}, finally, we manually create a validation subset by sampling ten per cent of each original training partition for each MIT Corpus following previous methodologies \cite{raman2022transforming, farina2023distillation}. Table \ref{tab:corpora_stats} shows the statistics for the corpora we used in this work. It is important to mention that although there are some duplicates in the datasets, we do not remove these instances.

\begin{table}[!ht]
    \centering
    \tiny
    \resizebox{\columnwidth}{!}{%
    \renewcommand{\arraystretch}{0.80}
\begin{tabular}{lcccc}
        \toprule
        \textbf{Corpora} & \textbf{Train} & \textbf{Val} & \textbf{Test} & \textbf{Labels}\\
        \midrule
        ATIS & 4,478 & 500 & 893 & 83 \\
        SNIPS & 13,084 & 700 & 700 & 39 \\
        MovieTrivia & 7,034 & 782 & 1,953 & 12 \\
        Movie & 8,797 & 978 & 2,443 & 12 \\
        Restaurant & 6,894 & 766 & 1,521 & 8 \\
        mTOP & 15,667 & 2,235 & 4,386 & 75 \\
        mTOD & 30,521 & 4,181 & 8,621 & 9 \\
        \midrule
    \end{tabular}}
    \caption{Statistics per partition of the used datasets.}
    \label{tab:corpora_stats}
\end{table}

\subsection{Experimental Details and Evaluation}\label{subsec:experimental_setings}
We conduct experiments to compare our model with other parameter-efficient fine-tuning methods. Our first goal is to assess the advantages of combining different parameter-efficient techniques in a multi-task setting versus a single-task setting. The second goal is to evaluate our model against other hypernetwork-based approaches in a multi-task environment.

We compare our approaches with the UniPELT framework \cite{mao-etal-2022-unipelt}, HyperFormer model \cite{karimi-mahabadi-etal-2021-parameter} and Hyperdecoder model \cite{ivison-peters-2022-hyperdecoders} described in section \ref{sec:related work}. We also compare our approach using Adapters and LoRA with T5 as the backbone model in a single-task setting. We found that the prefix-tuning and the gating mechanism proposed in the UniPELT framework decrease the performance using T5. Therefore, we only use Adapters and LoRA without a gating mechanism for our baseline comparison for the combination of UniPELT and T5. For all the implementations, we use the HuggingFace's Transformers \cite{wolf-etal-2020-transformers} and AdapterHub \cite{poth-etal-2023-adapters} libraries.

We use the batch sizes, learning rates and epochs reported in the original paper of each baseline to run the experiments in our work. For our proposed HyperLoader model, we used the same codebase\footnote{\url{https://github.com/rabeehk/hyperformer}} (under the Apache License, Version 2.0) and almost the same experimental details described by \cite{karimi-mahabadi-etal-2021-parameter}, that is, we employ the T5 base model, a batch size of 32, a constant learning rate of $3 \times 10^{-4}$, $2^{18}$ steps in all experiments, save a checkpoint every 1,000 steps, and sample tasks with conventional temperature-based with temperature $T=10$ proportional to $p^{1/T}_{\tau}$, where $p_{\tau}=\frac{N_\tau}{\sum_{i=1}^{T}N_{\tau}}$ and $N_{\tau}$ is the number of training samples for the $\tau$-th task. The only difference in our work is that we choose the best model based on the loss value instead of the average evaluation metric performance. We performed the experiments on 4 Nvidia A100 SXM 40G GPUs.    

%\subsection{Evaluation}
We use the micro-averaged F1-score as the evaluation metric for all experiments following the CoNLL convention \cite{tjong-kim-sang-de-meulder-2003-introduction}, where an entity is considered correct only if the entity is predicted exactly as it appears in the gold data. We use the SeqEval framework \cite{seqeval} to compute the scores in all the experiments.

\section{Results}
This section presents the results of the experiments we performed in this work described in section \ref{subsec:experimental_setings}. 
\subsection{Full Dataset Performance}
Table \ref{subtab:full_results} shows the performance of our proposed approach and the baselines using the 100\% of training and evaluation data. The UniPELT framework, Adapters and LoRA follow a single-task fine-tuning approach, therefore, the trainable parameters correspond to each dataset considered. For UniPELT, we report the results using BERT as in the original paper and T5 to compare the performance depending on the model's architecture. Initial experiments show that the use of the Prefix-tuning method and the gating mechanism decreases the performance of the encoder-decoder model; for this reason, we do not use these elements when applying UniPELT to T5. 

Based on the results in Table \ref{subtab:full_results}, it is possible to observe that our proposed HyperLoader model outperforms the baselines in three of the seven datasets and obtains the highest average with 0.8811, followed by the HyperFormer model (0.8795), UniPELT with T5 (0.8740), T5 with Adapters (0.8735), T5 with LoRA (0.8717), the Hyperdecoder model (0.8709) and UniPELT with BERT (0.8537). In terms of single-task and multi-task settings, it is possible to observe that the combination of UNiPELT with T5 outperforms the other models in two datasets (Movie with 0.8839, and ATIS with 0.9611); this result demonstrates that the use of hypernetworks improves the sharing information capacity of the models while reducing the negative transfer between tasks. Finally, when comparing the multi-task approaches based on hypernetworks, HyperLoader obtained a better average result than the HyperFormer model, which is the primary foundation of our work. In addition, since Hyperdecoder has more trainable parameters, the results indicate that the superior performance of our model is not only due to having more trainable parameters than the HyperFormer approach but also due to the combination of different parameter-efficient methods in a multi-task setting.     

\begin{table*}[!ht]
    \begin{subtable}[h]{\textwidth}
    \centering
    \scriptsize
    \resizebox{\textwidth}{!}{%
    \renewcommand{\arraystretch}{0.70}
    \begin{tabular}{l|ccccccc}
      \toprule
      & \textbf{Adapters} & \textbf{LoRA} & \textbf{UniPELT} & \textbf{UniPELT} & \textbf{HyperFormer} & \textbf{HyperLoader} & \textbf{Hyperdecoder} \\
      \textbf{Dataset} & \scriptsize T5[222M/3.5M] $\clubsuit$ & \scriptsize T5[222M/0.84M] $\clubsuit$ & \scriptsize BERT[110M/1.8M] $\clubsuit$ & \scriptsize T5[220M/4.5M] $\clubsuit$& \scriptsize T5[228M/5M] & \scriptsize T5[229M/6M] & \scriptsize T5[239M/16M]\\
      \midrule
      MovieTriva & 0.7184 & 0.7183 & 0.6873 & 0.7210 & \textbf{0.7295} & 0.7140 & 0.7072 \\
      Movie & 0.8793 & 0.8709 & 0.8500 & \textbf{0.8839} & 0.8788 & 0.8745 & 0.8700 \\
      Restaurant & 0.8069 & 0.8186 & 0.7689 & 0.8039 & \textbf{0.8136} & 0.7965 & 0.7891 \\
      ATIS & 0.9581 & 0.9549 & 0.9446 & \textbf{0.9611} & 0.9595 & 0.9599 & 0.9519 \\
      SNIPS & 0.9573 & 0.9506 & 0.9238 & 0.9478 & 0.9462 & \textbf{0.9592} & 0.9377 \\
      mTOP & 0.8299 & 0.8254 & 0.8432 & 0.8374 & 0.8638 & \textbf{0.8983} & 0.8752 \\
      mTOD & 0.9647 & 0.9633 & 0.9580 & 0.9627 & 0.9648 & \textbf{0.9652} & 0.9649 \\  
      \midrule
      \textbf{Average} & 0.8735 & 0.8717 & 0.8537 & 0.8768 & 0.8795 & \textbf{0.8811} & 0.8709 \\   
      \end{tabular}}
      \caption{Full dataset performance}
      \label{subtab:full_results}
      \end{subtable}
    \begin{subtable}[h]{\textwidth}
    \centering
    \scriptsize
    \resizebox{\textwidth}{!}{%
    \renewcommand{\arraystretch}{0.70}
    \begin{tabular}{l|ccccccc}
      \toprule
      & \textbf{Adapters} & \textbf{LoRA} & \textbf{UniPELT} & \textbf{UniPELT} & \textbf{HyperFormer} & \textbf{HyperLoader} & \textbf{Hyperdecoder} \\
      \textbf{Dataset} & \scriptsize T5[222M/3.5M] $\clubsuit$ & \scriptsize T5[222M/0.84M] $\clubsuit$ & \scriptsize BERT[110M/1.8M] $\clubsuit$ & \scriptsize T5[220M/4.5M] $\clubsuit$& \scriptsize T5[228M/5M] & \scriptsize T5[229M/6M] & \scriptsize T5[239M/16M]\\
      \midrule
      \rowcolor{gray!20}\multicolumn{8}{c}{10\% of Training and validation data}\\
      \midrule
      MovieTriva & 0.6633 & 0.6695 & 0.5152 & 0.6717 & 0.6589 & \textbf{0.6695} & 0.6597 \\
      Movie & 0.8281 & 0.8134 & 0.7809 & 0.8244 & 0.8311 & \textbf{0.8368} & 0.8202 \\
      Restaurant & 0.7292 & 0.7171&  0.5907 & 0.7049 & 0.7389 & \textbf{0.7499} & 0.7216 \\
      ATIS & 0.9129 & 0.8329 & 0.5901 & 0.8919 & 0.9162 & 0.9172 & \textbf{0.9193} \\
      SNIPS & 0.8647 & 0.8460 & 0.7398 & 0.8991 & 0.8943 & \textbf{0.8970} & 0.8727 \\
      mTOP & 0.7559 & 0.6886 & 0.6048 & 0.7467 & 0.7976 & \textbf{0.8114} & 0.7760 \\
      mTOD & 0.9459 & 0.9453 & 0.9351 & 0.9436 & 0.9467 & \textbf{0.9500} & 0.9436 \\
      \midrule
      \textbf{Average} & 0.8143 & 0.7875 & 0.6795 & 0.8117 & 0.8262 & \textbf{0.8331} & 0.8162 \\
      \midrule
      \rowcolor{gray!20}\multicolumn{8}{c}{20\% of Training and validation data}\\
      \midrule
      MovieTriva & 0.6896 & 0.6934 & 0.6105 & \textbf{0.7036} & 0.6729 & 0.6863 & 0.6767 \\
      Movie & 0.8492 & 0.8276 & 0.8047 & 0.8444 & 0.8407 & \textbf{0.8546} & 0.8377 \\
      Restaurant & \textbf{0.7699} & 0.7522 & 0.7114 & 0.7562 & 0.7517 & 0.7601 & 0.7401 \\
      ATIS & 0.9295 & 0.9222 & 0.8048 & 0.9281 & 0.9382 & \textbf{0.9401} & 0.9311 \\
      SNIPS & 0.9250 & 0.9096 & 0.8474 & 0.9235 & \textbf{0.9252} & 0.9238 & 0.9044 \\
      mTOP & 0.7798 & 0.7597 & 0.7052 & 0.7863 & 0.8399 & \textbf{0.8474} & 0.8140 \\
      mTOD & 0.9424 & 0.9543 & 0.9433 & 0.9522 & \textbf{0.9543} & 0.9541 & 0.9529  \\
      \midrule
      \textbf{Average} & 0.8408 & 0.8313 & 0.7753 & 0.8420 & 0.8461 & \textbf{0.8523} & 0.8367 \\
   \end{tabular}}
      \caption{Low-resource setting results}
      \label{subtab:low_resource}
      \end{subtable}
    \caption{Performance of the HyperLoader model and baselines is shown for 100\% (a), 10\%, and 20\% (b) of the training and evaluation data. Each method includes the model used and the total/trainable parameters in brackets. $\clubsuit$ denotes a single-task fine-tuning approach, so trainable parameters are specific to each dataset. Bold fonts highlight the best result for each dataset and on average. Micro-averaged F1-score is reported.}
    \label{tab:results}
\end{table*}
\subsection{Low-resource Setting Results}
We evaluate the performance of our proposed model and the baselines in a low-resource setting to measure its robustness with a data scarcity constraint, a common scenario in NLP \cite{hedderich-etal-2021-survey}. For this purpose, we down-sampled the training and validation partitions of all datasets to 10\% and 20\% of their original size and evaluated them in the full test subset. Table \ref{subtab:low_resource} shows the results of these experiments. 

The first part of Table \ref{subtab:low_resource} contains the performance of the HyperLoader model and the baselines when only 10\% of the training and validation data is used. It is evident that the combination of the UniPELT framework and the BERT model experiences the most significant performance decrease, from an average of 0.8537 to 0.6795. In contrast, the T5-based approaches show a smaller performance decrease in both single-task and multi-task settings. These results suggest that an encoder-decoder model is more capable of handling low-resource settings in both single-task and multi-task configurations. For the multi-task approaches, our HyperLoader model outperforms the HyperFormer and Hyperdecoder approaches in six out of seven datasets and on the average performance with a score of 0.8331, followed by HyperFormer (0.8262) and Hyperdecoder (0.8161). 

\begin{figure}[!ht]
    \centering
    \includegraphics[scale=0.225]{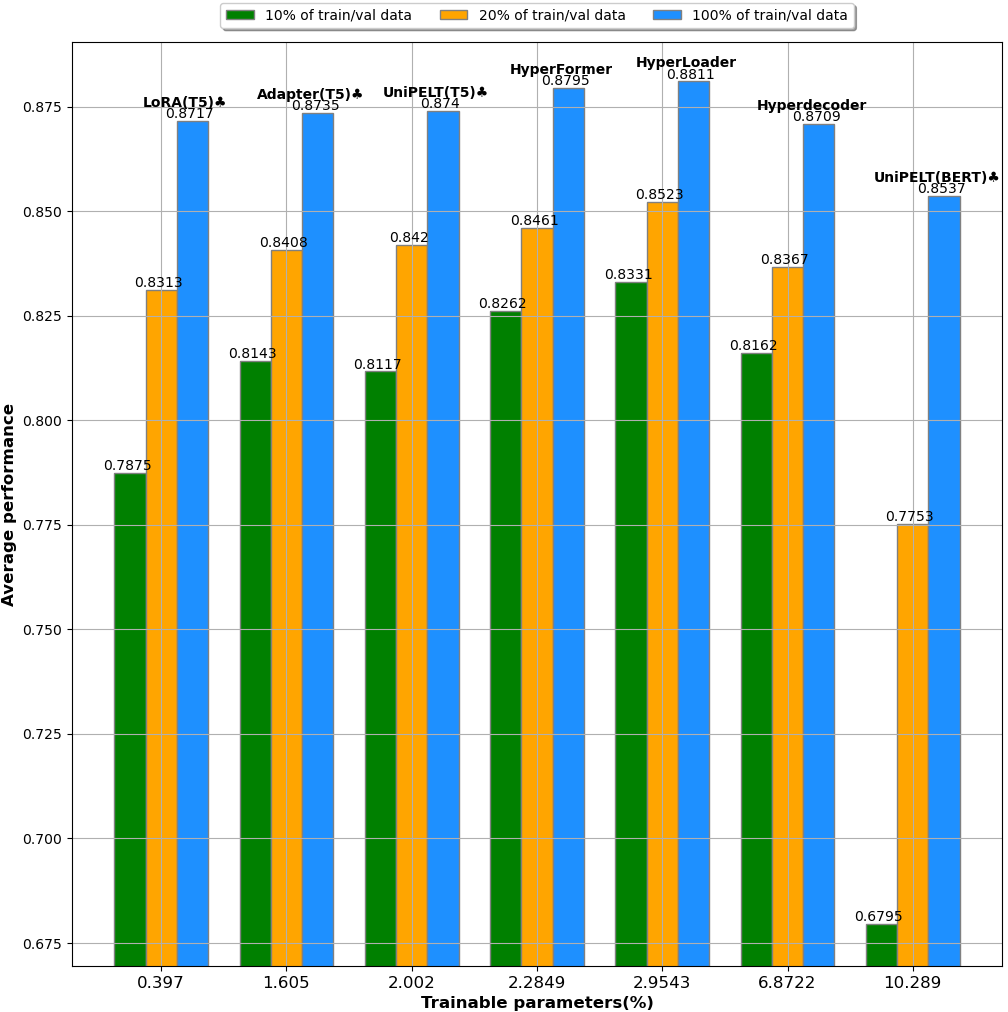}
    \caption{Average performance-percentage of trainable parameters plot using different portions of the datasets. $\clubsuit$ indicates a single-task fine-tuning approach.}
    \label{fig:percentage_performance}
\end{figure}

The second part of Table \ref{subtab:low_resource} shows the results when 20\% of the training and evaluation data is used. Using this amount of data our HyperLoader model outperforms the baselines in three of seven datasets and on the average performance. It is also noticeable that the combination of UniPELT and T5 using a single-task fine-tuning approach outperforms the other models only in the MovieTrivia dataset, with a micro-averaged F1-score of 0.7036 followed by HyperLoader with 0.6863, Hyperdecoder with 0.6767, HyperFormer with 0.6729 and finally UniPELT using BERT with 0.6105. These results suggest that combining various parameter-efficient methods in a multi-task setting and generating their weights based on the task, transformer layer, and the position where the parameter-efficient method is applied (Adapter layer or LoRA matrix) is effective for Sequence Labelling tasks, leading to good performance with a parameter-efficient approach.
Figure \ref{fig:percentage_performance} shows the relationship between average performance and the percentage of trainable parameters per model. The symbol $\clubsuit$ denotes a single fine-tuning approach. Our approach consistently outperforms other methods across 10\%, 20\%, and 100\% of the training and validation data. This evidence shows that combining various parameter-efficient fine-tuning techniques in a multi-task setting is effective for low-resource scenarios. The performance boost is not just due to adding more parameters, as seen with the hyperdecoder model and the combination of UniPELT with BERT, which have a higher percentage of trainable parameters. Therefore, the combination of parameter-efficient methods and weight generation is a potent alternative for Sequence Labelling tasks. 

\subsection{Ablation Experiments}
We conducted ablation experiments to assess the effect of varying the number of trainable parameters in our model. Table \ref{tab:ablation} presents the results in descending order of performance. The first row (in blue) corresponds to the final version of our model, as described in section \ref{sec:methodology}. The fifth row (in grey) corresponds to the HyperFormer model, which is the main foundation of our work.

The initial experiments (denoted by +) involved increasing the number of trainable parameters for generating LoRA's weight matrices. In the case of using four hypernetworks, the first one generates the Low-rank A matrix for Query (Q) and Value (V) in the self-attention module, and the second generates the Low-rank B matrix for Q and V in the self-attention module. Analogously, the third and fourth hypernetworks generate the A and B matrices in the cross-attention module of the decoder. In the case of 8 hypernetworks, each generates the weight of only one component: one for the A and one for the B matrices in the Query; another two hypernetworks for the A and B matrices in the Value of the self-attention. The remaining four hypernetworks are analogous to the former but in the cross-attention component of the decoder. Finally, the last experiment involves increasing the reduction factor ($r=16$) of the LoRA matrices in the final version of the HyperLoader model. These results indicate that our model achieves a strong balance between performance and the percentage of trainable parameters (2.9543\%), with only a 0.6694\% increase compared to the HyperFormer model (2.3849\%). It also shows that using only 2 hypernetworks for the LoRA matrices enables the model to generate better weights based on the task, the parameter-efficient method and its position within the Transformer layer of the model.

The second experiment involved removing (denoted by $-$) the Adapters from the model and adjusting only the task-conditioned LoRA matrices during training. This version showed the lowest average performance, though the difference from the HyperFormer model was minimal (0.0001). This suggests that while a single parameter-efficient technique can effectively address different tasks, combining such methods provides better results in a multi-task setting and shows that further research into this area is necessary.
\begin{table}%[!ht]
    \centering
    \resizebox{\columnwidth}{!}{%
    \renewcommand{\arraystretch}{0.80}
    \begin{tabular}{lcc}
        \toprule
        \textbf{Variant} & \textbf{Total/trainable params} & \textbf{Avg score}\\
        \midrule
        \rowcolor{cyan!20}HyperLoader & 229M/6M & \textbf{0.8811} \\
        + 4 LoRA HNs & 228M/8M & 0.0808\\
        + 8 LoRA HNs & 234M/11M & 0.800\\
        + 2 LoRA HNs, r=16 & 229M/6M & 0.8804 \\
        \rowcolor{gray!20}HyperFormer & 228M/5M & 0.8795 \\
        - Adapters & 225M/1.8M & 0.8794 \\
    \end{tabular}}
    \caption{Average performance of the ablation experiments. We denote hypernetworks as HNs.}
    \label{tab:ablation}
\end{table}

\section{Conclusions}
We proposed a method that combines parameter-efficient methods in a multi-task setting. Using a hypernetwork, we generate the weights of the adapter layers and LoRA matrices conditioned on the task, the transformer layer of the model, and the position of these modules within this layer. Our method outperforms previous work in single-task and multi-task fine-tuning that combines different parameter-efficient methods and hypernetwork-based approaches using the full training and validation splits of each dataset and in a low-resource configuration using only 10\% and 20\% of those partitions to train the model. We provide empirical evidence that the improvement in performance is not only because of augmenting the trainable parameters since the hyperdecoder model has the largest number of trainable weights. Therefore, the combination of parameter-efficient methods and weight generation is a strong alternative to solving Sequence Labelling tasks in a multi-task setting.

\section{Limitations and Future work}
Our model excels in multi-task settings with a parameter-efficient fine-tuning approach that mitigates negative transfer, underfitting, and overfitting. However, it still requires access to all datasets during training and needs complete retraining when a new task is added. Curriculum learning \cite{bengio2009curriculum, wang2021survey, soviany2022curriculum, piergiovannidynamic} could address this limitation by enhancing learning efficiency, potentially leading to faster convergence and improved performance.
\bibliography{acl_latex}

\begin{thebibliography}{38}
\providecommand{\natexlab}[1]{#1}

\bibitem[{Bengio et~al.(2009)Bengio, Louradour, Collobert, and Weston}]{bengio2009curriculum}
Yoshua Bengio, J\'{e}r\^{o}me Louradour, Ronan Collobert, and Jason Weston. 2009.
\newblock \href {https://doi.org/10.1145/1553374.1553380} {Curriculum learning}.
\newblock In \emph{Proceedings of the 26th Annual International Conference on Machine Learning}, ICML '09, page 41–48, New York, NY, USA. Association for Computing Machinery.

\bibitem[{Cheng et~al.(2023{\natexlab{a}})Cheng, Zhu, Cao, Ye, and Zou}]{cheng-etal-2023-mrrl}
Xuxin Cheng, Zhihong Zhu, Bowen Cao, Qichen Ye, and Yuexian Zou. 2023{\natexlab{a}}.
\newblock \href {https://doi.org/10.18653/v1/2023.findings-emnlp.704} {{MRRL}: Modifying the reference via reinforcement learning for non-autoregressive joint multiple intent detection and slot filling}.
\newblock In \emph{Findings of the Association for Computational Linguistics: EMNLP 2023}, pages 10495--10505, Singapore. Association for Computational Linguistics.

\bibitem[{Cheng et~al.(2023{\natexlab{b}})Cheng, Zhu, Xu, Li, Li, and Zou}]{cheng-etal-2023-accelerating}
Xuxin Cheng, Zhihong Zhu, Wanshi Xu, Yaowei Li, Hongxiang Li, and Yuexian Zou. 2023{\natexlab{b}}.
\newblock \href {https://doi.org/10.18653/v1/2023.findings-emnlp.597} {Accelerating multiple intent detection and slot filling via targeted knowledge distillation}.
\newblock In \emph{Findings of the Association for Computational Linguistics: EMNLP 2023}, pages 8900--8910, Singapore. Association for Computational Linguistics.

\bibitem[{Coucke et~al.(2018)Coucke, Saade, Ball, Bluche, Caulier, Leroy, Doumouro, Gisselbrecht, Caltagirone, Lavril et~al.}]{coucke2018snips}
Alice Coucke, Alaa Saade, Adrien Ball, Th{\'e}odore Bluche, Alexandre Caulier, David Leroy, Cl{\'e}ment Doumouro, Thibault Gisselbrecht, Francesco Caltagirone, Thibaut Lavril, et~al. 2018.
\newblock Snips voice platform: an embedded spoken language understanding system for private-by-design voice interfaces.
\newblock \emph{arXiv preprint arXiv:1805.10190}.

\bibitem[{Farina et~al.(2023)Farina, Pappadopulo, Gupta, Huang, {\.I}rsoy, and Solorio}]{farina2023distillation}
Marco Farina, Duccio Pappadopulo, Anant Gupta, Leslie Huang, Ozan {\.I}rsoy, and Thamar Solorio. 2023.
\newblock Distillation of encoder-decoder transformers for sequence labelling.
\newblock In \emph{Findings of the Association for Computational Linguistics: EACL 2023}, pages 2539--2549.

\bibitem[{Firdaus et~al.(2023)Firdaus, Ekbal, and Cambria}]{firdaus2023multitask}
Mauajama Firdaus, Asif Ekbal, and Erik Cambria. 2023.
\newblock Multitask learning for multilingual intent detection and slot filling in dialogue systems.
\newblock \emph{Information Fusion}, 91:299--315.

\bibitem[{Goo et~al.(2018)Goo, Gao, Hsu, Huo, Chen, Hsu, and Chen}]{goo-etal-2018-slot}
Chih-Wen Goo, Guang Gao, Yun-Kai Hsu, Chih-Li Huo, Tsung-Chieh Chen, Keng-Wei Hsu, and Yun-Nung Chen. 2018.
\newblock \href {https://doi.org/10.18653/v1/N18-2118} {Slot-gated modeling for joint slot filling and intent prediction}.
\newblock In \emph{Proceedings of the 2018 Conference of the North {A}merican Chapter of the Association for Computational Linguistics: Human Language Technologies, Volume 2 (Short Papers)}, pages 753--757, New Orleans, Louisiana. Association for Computational Linguistics.

\bibitem[{Ha et~al.(2017)Ha, Dai, and Le}]{ha2016hypernetworks}
David Ha, Andrew~M. Dai, and Quoc~V. Le. 2017.
\newblock \href {https://openreview.net/forum?id=rkpACe1lx} {Hypernetworks}.
\newblock In \emph{5th International Conference on Learning Representations, {ICLR} 2017, Toulon, France, April 24-26, 2017, Conference Track Proceedings}. OpenReview.net.

\bibitem[{Hedderich et~al.(2021)Hedderich, Lange, Adel, Str{\"o}tgen, and Klakow}]{hedderich-etal-2021-survey}
Michael~A. Hedderich, Lukas Lange, Heike Adel, Jannik Str{\"o}tgen, and Dietrich Klakow. 2021.
\newblock \href {https://doi.org/10.18653/v1/2021.naacl-main.201} {A survey on recent approaches for natural language processing in low-resource scenarios}.
\newblock In \emph{Proceedings of the 2021 Conference of the North American Chapter of the Association for Computational Linguistics: Human Language Technologies}, pages 2545--2568, Online. Association for Computational Linguistics.

\bibitem[{Hemphill et~al.(1990)Hemphill, Godfrey, and Doddington}]{hemphill-etal-1990-atis}
Charles~T. Hemphill, John~J. Godfrey, and George~R. Doddington. 1990.
\newblock \href {https://aclanthology.org/H90-1021} {The {ATIS} spoken language systems pilot corpus}.
\newblock In \emph{Speech and Natural Language: Proceedings of a Workshop Held at Hidden Valley, {P}ennsylvania, June 24-27,1990}.

\bibitem[{Houlsby et~al.(2019)Houlsby, Giurgiu, Jastrzebski, Morrone, De~Laroussilhe, Gesmundo, Attariyan, and Gelly}]{houlsby19a}
Neil Houlsby, Andrei Giurgiu, Stanislaw Jastrzebski, Bruna Morrone, Quentin De~Laroussilhe, Andrea Gesmundo, Mona Attariyan, and Sylvain Gelly. 2019.
\newblock \href {https://proceedings.mlr.press/v97/houlsby19a.html} {Parameter-efficient transfer learning for {NLP}}.
\newblock In \emph{Proceedings of the 36th International Conference on Machine Learning}, volume~97 of \emph{Proceedings of Machine Learning Research}, pages 2790--2799. PMLR.

\bibitem[{Hu et~al.(2021)Hu, Shen, Wallis, Allen-Zhu, Li, Wang, Wang, and Chen}]{hu2021lora}
Edward~J Hu, Yelong Shen, Phillip Wallis, Zeyuan Allen-Zhu, Yuanzhi Li, Shean Wang, Lu~Wang, and Weizhu Chen. 2021.
\newblock Lora: Low-rank adaptation of large language models.
\newblock \emph{arXiv preprint arXiv:2106.09685}.

\bibitem[{Hu et~al.(2024)Hu, Chen, Du, Peng, Keloth, Zuo, Zhou, Li, Jiang, Lu, Roberts, and Xu}]{10.1093/jamia/ocad259}
Yan Hu, Qingyu Chen, Jingcheng Du, Xueqing Peng, Vipina~Kuttichi Keloth, Xu~Zuo, Yujia Zhou, Zehan Li, Xiaoqian Jiang, Zhiyong Lu, Kirk Roberts, and Hua Xu. 2024.
\newblock \href {https://doi.org/10.1093/jamia/ocad259} {{Improving large language models for clinical named entity recognition via prompt engineering}}.
\newblock \emph{Journal of the American Medical Informatics Association}, page ocad259.

\bibitem[{Ivison and Peters(2022)}]{ivison-peters-2022-hyperdecoders}
Hamish Ivison and Matthew Peters. 2022.
\newblock \href {https://doi.org/10.18653/v1/2022.findings-emnlp.124} {Hyperdecoders: Instance-specific decoders for multi-task {NLP}}.
\newblock In \emph{Findings of the Association for Computational Linguistics: EMNLP 2022}, pages 1715--1730, Abu Dhabi, United Arab Emirates. Association for Computational Linguistics.

\bibitem[{Karimi~Mahabadi et~al.(2021{\natexlab{a}})Karimi~Mahabadi, Henderson, and Ruder}]{NEURIPS2021_081be9fd}
Rabeeh Karimi~Mahabadi, James Henderson, and Sebastian Ruder. 2021{\natexlab{a}}.
\newblock \href {https://proceedings.neurips.cc/paper_files/paper/2021/file/081be9fdff07f3bc808f935906ef70c0-Paper.pdf} {Compacter: Efficient low-rank hypercomplex adapter layers}.
\newblock In \emph{Advances in Neural Information Processing Systems}, volume~34, pages 1022--1035. Curran Associates, Inc.

\bibitem[{Karimi~Mahabadi et~al.(2021{\natexlab{b}})Karimi~Mahabadi, Ruder, Dehghani, and Henderson}]{karimi-mahabadi-etal-2021-parameter}
Rabeeh Karimi~Mahabadi, Sebastian Ruder, Mostafa Dehghani, and James Henderson. 2021{\natexlab{b}}.
\newblock \href {https://doi.org/10.18653/v1/2021.acl-long.47} {Parameter-efficient multi-task fine-tuning for transformers via shared hypernetworks}.
\newblock In \emph{Proceedings of the 59th Annual Meeting of the Association for Computational Linguistics and the 11th International Joint Conference on Natural Language Processing (Volume 1: Long Papers)}, pages 565--576, Online. Association for Computational Linguistics.

\bibitem[{Lee et~al.(2017)Lee, Cho, and Hofmann}]{lee-etal-2017-fully}
Jason Lee, Kyunghyun Cho, and Thomas Hofmann. 2017.
\newblock \href {https://doi.org/10.1162/tacl_a_00067} {Fully character-level neural machine translation without explicit segmentation}.
\newblock \emph{Transactions of the Association for Computational Linguistics}, 5:365--378.

\bibitem[{Li et~al.(2021)Li, Arora, Chen, Gupta, Gupta, and Mehdad}]{li-etal-2021-mtop}
Haoran Li, Abhinav Arora, Shuohui Chen, Anchit Gupta, Sonal Gupta, and Yashar Mehdad. 2021.
\newblock \href {https://doi.org/10.18653/v1/2021.eacl-main.257} {{MTOP}: A comprehensive multilingual task-oriented semantic parsing benchmark}.
\newblock In \emph{Proceedings of the 16th Conference of the European Chapter of the Association for Computational Linguistics: Main Volume}, pages 2950--2962, Online. Association for Computational Linguistics.

\bibitem[{Li and Liang(2021)}]{li-liang-2021-prefix}
Xiang~Lisa Li and Percy Liang. 2021.
\newblock \href {https://doi.org/10.18653/v1/2021.acl-long.353} {Prefix-tuning: Optimizing continuous prompts for generation}.
\newblock In \emph{Proceedings of the 59th Annual Meeting of the Association for Computational Linguistics and the 11th International Joint Conference on Natural Language Processing (Volume 1: Long Papers)}, pages 4582--4597, Online. Association for Computational Linguistics.

\bibitem[{Mao et~al.(2022)Mao, Mathias, Hou, Almahairi, Ma, Han, Yih, and Khabsa}]{mao-etal-2022-unipelt}
Yuning Mao, Lambert Mathias, Rui Hou, Amjad Almahairi, Hao Ma, Jiawei Han, Scott Yih, and Madian Khabsa. 2022.
\newblock \href {https://doi.org/10.18653/v1/2022.acl-long.433} {{U}ni{PELT}: A unified framework for parameter-efficient language model tuning}.
\newblock In \emph{Proceedings of the 60th Annual Meeting of the Association for Computational Linguistics (Volume 1: Long Papers)}, pages 6253--6264, Dublin, Ireland. Association for Computational Linguistics.

\bibitem[{Nakayama(2018)}]{seqeval}
Hiroki Nakayama. 2018.
\newblock \href {https://github.com/chakki-works/seqeval} {{seqeval}: A python framework for sequence labeling evaluation}.
\newblock Software available from https://github.com/chakki-works/seqeval.

\bibitem[{Pfeiffer et~al.(2021)Pfeiffer, Kamath, R{\"u}ckl{\'e}, Cho, and Gurevych}]{pfeiffer-etal-2021-adapterfusion}
Jonas Pfeiffer, Aishwarya Kamath, Andreas R{\"u}ckl{\'e}, Kyunghyun Cho, and Iryna Gurevych. 2021.
\newblock \href {https://doi.org/10.18653/v1/2021.eacl-main.39} {{A}dapter{F}usion: Non-destructive task composition for transfer learning}.
\newblock In \emph{Proceedings of the 16th Conference of the European Chapter of the Association for Computational Linguistics: Main Volume}, pages 487--503, Online. Association for Computational Linguistics.

\bibitem[{Pfeiffer et~al.(2020)Pfeiffer, Vuli{\'c}, Gurevych, and Ruder}]{pfeiffer-etal-2020-mad}
Jonas Pfeiffer, Ivan Vuli{\'c}, Iryna Gurevych, and Sebastian Ruder. 2020.
\newblock \href {https://doi.org/10.18653/v1/2020.emnlp-main.617} {{MAD-X}: {A}n {A}dapter-{B}ased {F}ramework for {M}ulti-{T}ask {C}ross-{L}ingual {T}ransfer}.
\newblock In \emph{Proceedings of the 2020 Conference on Empirical Methods in Natural Language Processing (EMNLP)}, pages 7654--7673, Online. Association for Computational Linguistics.

\bibitem[{Piergiovanni et~al.(2023)Piergiovanni, Kuo, Li, and Angelova}]{piergiovannidynamic}
AJ~Piergiovanni, Weicheng Kuo, Wei Li, and Anelia Angelova. 2023.
\newblock Dynamic pre-training of vision-language models.
\newblock In \emph{Workshop on Multimodal Representation Learning. ICLR 2023}.

\bibitem[{Poth et~al.(2023)Poth, Sterz, Paul, Purkayastha, Engl{\"a}nder, Imhof, Vuli{\'c}, Ruder, Gurevych, and Pfeiffer}]{poth-etal-2023-adapters}
Clifton Poth, Hannah Sterz, Indraneil Paul, Sukannya Purkayastha, Leon Engl{\"a}nder, Timo Imhof, Ivan Vuli{\'c}, Sebastian Ruder, Iryna Gurevych, and Jonas Pfeiffer. 2023.
\newblock \href {https://aclanthology.org/2023.emnlp-demo.13} {Adapters: A unified library for parameter-efficient and modular transfer learning}.
\newblock In \emph{Proceedings of the 2023 Conference on Empirical Methods in Natural Language Processing: System Demonstrations}, pages 149--160, Singapore. Association for Computational Linguistics.

\bibitem[{Raffel et~al.(2020)Raffel, Shazeer, Roberts, Lee, Narang, Matena, Zhou, Li, and Liu}]{raffel2020exploring}
Colin Raffel, Noam Shazeer, Adam Roberts, Katherine Lee, Sharan Narang, Michael Matena, Yanqi Zhou, Wei Li, and Peter~J Liu. 2020.
\newblock Exploring the limits of transfer learning with a unified text-to-text transformer.
\newblock \emph{Journal of machine learning research}, 21(140):1--67.

\bibitem[{Raman et~al.(2022)Raman, Naim, Chen, Hashimoto, Yalasangi, and Srinivasan}]{raman2022transforming}
Karthik Raman, Iftekhar Naim, Jiecao Chen, Kazuma Hashimoto, Kiran Yalasangi, and Krishna Srinivasan. 2022.
\newblock Transforming sequence tagging into a seq2seq task.
\newblock In \emph{Proceedings of the 2022 Conference on Empirical Methods in Natural Language Processing}, pages 11856--11874.

\bibitem[{Raza et~al.(2022)Raza, Reji, Shajan, and Bashir}]{raza2022large}
Shaina Raza, Deepak~John Reji, Femi Shajan, and Syed~Raza Bashir. 2022.
\newblock Large-scale application of named entity recognition to biomedicine and epidemiology.
\newblock \emph{PLOS Digital Health}, 1(12):e0000152.

\bibitem[{R{\"u}ckl{\'e} et~al.(2021)R{\"u}ckl{\'e}, Geigle, Glockner, Beck, Pfeiffer, Reimers, and Gurevych}]{ruckle-etal-2021-adapterdrop}
Andreas R{\"u}ckl{\'e}, Gregor Geigle, Max Glockner, Tilman Beck, Jonas Pfeiffer, Nils Reimers, and Iryna Gurevych. 2021.
\newblock \href {https://doi.org/10.18653/v1/2021.emnlp-main.626} {{AdapterDrop}: {O}n the efficiency of adapters in transformers}.
\newblock In \emph{Proceedings of the 2021 Conference on Empirical Methods in Natural Language Processing}, pages 7930--7946, Online and Punta Cana, Dominican Republic. Association for Computational Linguistics.

\bibitem[{Schuster et~al.(2019)Schuster, Gupta, Shah, and Lewis}]{schuster-etal-2019-cross-lingual}
Sebastian Schuster, Sonal Gupta, Rushin Shah, and Mike Lewis. 2019.
\newblock \href {https://doi.org/10.18653/v1/N19-1380} {Cross-lingual transfer learning for multilingual task oriented dialog}.
\newblock In \emph{Proceedings of the 2019 Conference of the North {A}merican Chapter of the Association for Computational Linguistics: Human Language Technologies, Volume 1 (Long and Short Papers)}, pages 3795--3805, Minneapolis, Minnesota. Association for Computational Linguistics.

\bibitem[{Soviany et~al.(2022)Soviany, Ionescu, Rota, and Sebe}]{soviany2022curriculum}
Petru Soviany, Radu~Tudor Ionescu, Paolo Rota, and Nicu Sebe. 2022.
\newblock Curriculum learning: A survey.
\newblock \emph{International Journal of Computer Vision}, 130(6):1526--1565.

\bibitem[{Stickland and Murray(2019)}]{stickland19a}
Asa~Cooper Stickland and Iain Murray. 2019.
\newblock \href {https://proceedings.mlr.press/v97/stickland19a.html} {{BERT} and {PAL}s: Projected attention layers for efficient adaptation in multi-task learning}.
\newblock In \emph{Proceedings of the 36th International Conference on Machine Learning}, volume~97 of \emph{Proceedings of Machine Learning Research}, pages 5986--5995. PMLR.

\bibitem[{Tjong Kim~Sang and De~Meulder(2003)}]{tjong-kim-sang-de-meulder-2003-introduction}
Erik~F. Tjong Kim~Sang and Fien De~Meulder. 2003.
\newblock \href {https://aclanthology.org/W03-0419} {Introduction to the {C}o{NLL}-2003 shared task: Language-independent named entity recognition}.
\newblock In \emph{Proceedings of the Seventh Conference on Natural Language Learning at {HLT}-{NAACL} 2003}, pages 142--147.

\bibitem[{Trias et~al.(2021)Trias, Wang, Jaume, and Idreos}]{trias-etal-2021-named}
Fernando Trias, Hongming Wang, Sylvain Jaume, and Stratos Idreos. 2021.
\newblock \href {https://doi.org/10.18653/v1/2021.nllp-1.18} {Named entity recognition in historic legal text: A transformer and state machine ensemble method}.
\newblock In \emph{Proceedings of the Natural Legal Language Processing Workshop 2021}, pages 172--179, Punta Cana, Dominican Republic. Association for Computational Linguistics.

\bibitem[{Wang et~al.(2021)Wang, Chen, and Zhu}]{wang2021survey}
Xin Wang, Yudong Chen, and Wenwu Zhu. 2021.
\newblock A survey on curriculum learning.
\newblock \emph{IEEE transactions on pattern analysis and machine intelligence}, 44(9):4555--4576.

\bibitem[{Wang et~al.(2019)Wang, Dai, Poczos, and Carbonell}]{Wang_2019_CVPR}
Zirui Wang, Zihang Dai, Barnabas Poczos, and Jaime Carbonell. 2019.
\newblock Characterizing and avoiding negative transfer.
\newblock In \emph{Proceedings of the IEEE/CVF Conference on Computer Vision and Pattern Recognition (CVPR)}.

\bibitem[{Wolf et~al.(2020)Wolf, Debut, Sanh, Chaumond, Delangue, Moi, Cistac, Rault, Louf, Funtowicz, Davison, Shleifer, von Platen, Ma, Jernite, Plu, Xu, Scao, Gugger, Drame, Lhoest, and Rush}]{wolf-etal-2020-transformers}
Thomas Wolf, Lysandre Debut, Victor Sanh, Julien Chaumond, Clement Delangue, Anthony Moi, Pierric Cistac, Tim Rault, Rémi Louf, Morgan Funtowicz, Joe Davison, Sam Shleifer, Patrick von Platen, Clara Ma, Yacine Jernite, Julien Plu, Canwen Xu, Teven~Le Scao, Sylvain Gugger, Mariama Drame, Quentin Lhoest, and Alexander~M. Rush. 2020.
\newblock \href {https://www.aclweb.org/anthology/2020.emnlp-demos.6} {Transformers: State-of-the-art natural language processing}.
\newblock In \emph{Proceedings of the 2020 Conference on Empirical Methods in Natural Language Processing: System Demonstrations}, pages 38--45, Online. Association for Computational Linguistics.

\bibitem[{Zhou et~al.(2023)Zhou, Zhang, Gu, Chen, and Poon}]{zhou2023universalner}
Wenxuan Zhou, Sheng Zhang, Yu~Gu, Muhao Chen, and Hoifung Poon. 2023.
\newblock Universalner: Targeted distillation from large language models for open named entity recognition.
\newblock In \emph{The Twelfth International Conference on Learning Representations}.

\end{thebibliography}

\appendix

\section{Corpora labels}\label{sec:appendix}

Table \ref{tab:label_list} shows each Sequence Labelling dataset's complete list of labels used to train and evaluate the HyperLoader model. Based on the labels' names, we calculate label overlap between the datasets. Nine different labels are shared across the datasets, representing 3.78\% of the total labels.

\begin{table*}
    \centering
    \small
    \begin{tabular}{llll}
    \hline
    \rowcolor{gray!20}\multicolumn{4}{c}{\textbf{ATIS labels}}\\
    \hline
    fromloc.airport\_name & arrive\_time.start\_time & flight\_number & cost\_relative \\
    connect & flight\_days & restriction\_code & depart\_date.date\_relative \\
    return\_date.month\_name & mod & arrive\_date.month\_name & city\_name \\
    depart\_date.day\_number & compartment & depart\_time.start\_time & airline\_name \\
    meal & depart\_date.month\_name & time\_relative & return\_date.today\_relative \\
    depart\_time.period\_mod & flight\_mod & airport\_name & stoploc.airport\_code \\
    depart\_date.year & fare\_basis\_code & today\_relative & airport\_code \\
    fromloc.state\_name & toloc.city\_name & economy & booking\_class \\
    arrive\_date.today\_relative & arrive\_date.date\_relative & toloc.airport\_code & fromloc.airport\_code \\
    day\_number & stoploc.city\_name & state\_code & month\_name \\
    arrive\_date.day\_name & arrive\_time.period\_of\_day & state\_name & aircraft\_code \\
    period\_of\_day & return\_time.period\_mod & day\_name & stoploc.state\_code \\
    toloc.state\_code & depart\_time.time\_relative & toloc.airport\_name & return\_date.date\_relative \\
    fromloc.city\_name & return\_date.day\_number & depart\_time.time & depart\_date.day\_name \\
    arrive\_time.time & meal\_code & or & class\_type \\
    return\_date.day\_name & time & toloc.state\_name & arrive\_date.day\_number \\
    days\_code & arrive\_time.period\_mod & arrive\_time.time\_relative & flight\_stop \\
    depart\_time.period\_of\_day & transport\_type & round\_trip & meal\_description \\
    fare\_amount & toloc.country\_name & arrive\_time.end\_time & depart\_time.end\_time \\
    flight & fromloc.state\_code & depart\_date.today\_relative & flight\_time \\
    airline\_code & return\_time.period\_of\_day & stoploc.airport\_name & \\
    \hline
    \rowcolor{gray!20}\multicolumn{4}{c}{\textbf{mTOP labels}}\\
    \hline
    school & contact\_related & recipes\_diet & news\_source \\
    todo & recipes\_unit\_nutrition & person\_reminded & attendee \\
    recipes\_source & date\_time & news\_topic & music\_album\_title \\
    life\_event & contact\_method & recipes\_time\_preparation & type\_contact \\
    news\_reference & similarity & name\_app & recipes\_qualifier\_nutrition \\
    recipes\_cooking\_method & timer\_name & contact\_removed & employer \\
    recipes\_excluded\_ingredient & method\_recipes & type\_relation & group \\
    news\_type & content\_exact & ordinal & news\_category \\
    recipes\_unit\_measurement & user\_attendee\_event & music\_playlist\_title & sender \\
    music\_provider\_name & recipes\_dish & location & amount \\
    music\_rewind\_time & music\_type & alarm\_name & weather\_temperature\_unit \\
    music\_album\_modifier & category\_event & education\_degree & recipes\_meal \\
    period & music\_artist\_name & music\_radio\_id & method\_timer \\
    recipes\_cuisine & phone\_number & music\_genre & weather\_attribute \\
    music\_track\_title & recipes\_included\_ingredient & attribute\_event & method\_retrieval\_reminder \\
    major & contact\_added & recipes\_rating & contact \\
    gender & age & recipient & job \\
    recipes\_type & recipes\_type\_nutrition & attendee\_event & recipes\_attribute \\
    music\_playlist\_modifier & title\_event & type\_content & \\
    \hline
    \rowcolor{gray!20}\multicolumn{4}{c}{\textbf{SNIPS labels}}\\
    \hline
    object\_location\_type & poi & state & genre \\
    object\_name & album & playlist & movie\_type \\
    movie\_name & artist & restaurant\_name & restaurant\_type \\
    spatial\_relation & timerange & object\_part\_of\_series\_type & current\_location \\
    served\_dish & city & object\_select & music\_item \\
    country & cuisine & geographic\_poi & sort \\
    condition\_temperature & object\_type & party\_size\_description & service \\
    track & entity\_name & party\_size\_number & rating\_value \\
    playlist\_owner & condition\_description & rating\_unit & location\_name \\
    year & facility & best\_rating & \\
    \hline
    \rowcolor{gray!20}\multicolumn{4}{c}{\textbf{Movie labels}}\\
    \hline
    genre & rating & year & plot \\
    ratings\_average & director & song & title \\
    trailer & review & actor & character \\
    \hline
    \rowcolor{gray!20}\multicolumn{4}{c}{\textbf{MovieTrivia labels}}\\
    \hline
    award & relationship & quote & genre \\
    character\_name & director & plot & year \\
    soundtrack & origin & actor & opinion \\
    \hline
    \rowcolor{gray!20}\multicolumn{4}{c}{\textbf{mTOD labels}}\\
    \hline
    demonstrative\_reference & datetime & weather & negation \\
    alarm & news & timer & reminder \\
    location & & & \\
    \hline
    \rowcolor{gray!20}\multicolumn{4}{c}{\textbf{Restaurant labels}}\\
    \hline
    restaurant\_name & cuisine & rating & price \\
    dish & hours & amenity & location \\
    \hline
    \end{tabular}
    \caption{List of labels for each used Sequence labelling dataset to evaluate our proposed approach.}
    \label{tab:label_list}
\end{table*}

\end{document}